\documentclass[11pt, a4paper, logo, singlecolumn, copyright]{style/gensyn}

\usepackage[colorinlistoftodos,textsize=tiny,textwidth=35pt]{todonotes}
\newcommand{\Comments}{1}
\newcommand{\mynote}[2]{\ifnum\Comments=1\textcolor{#1}{#2}\fi}
\newcommand{\mytodo}[2]{\ifnum\Comments=1\todo[linecolor=#1!80!black,backgroundcolor=#1,bordercolor=#1!80!black]{#2}\fi}

\ifnum\Comments=1
\paperwidth=\dimexpr \paperwidth + 50pt\relax
\oddsidemargin=\dimexpr\oddsidemargin + 25pt\relax
\evensidemargin=\dimexpr\evensidemargin + 25pt\relax
\marginparwidth=\dimexpr \marginparwidth + 25pt\relax
\fi

\usepackage{array}
\newcolumntype{P}{>{\centering\arraybackslash}p{2.5cm}}
\newcolumntype{M}{>{\centering\arraybackslash\footnotesize}m{.78cm}}
\newcolumntype{S}{>{\centering\arraybackslash\tiny}m{2cm}}

\newtcbtheorem[auto counter,number within=section]{obs}%
{Observation}{fonttitle=\bfseries\upshape, fontupper=\slshape,
	arc=0mm, colback=cyan!5!white,colframe=cyan!75!white}{theorem}

\newtcbtheorem[auto counter,number within=section]{ins}%
{Insight}{fonttitle=\bfseries\upshape, fontupper=\slshape,
	arc=0mm, colback=green!5!white,colframe=green!75!white}{theorem}

\usepackage{amsmath,amsfonts,bm}

\def\eqref#1{equation~\ref{#1}}

\def\1{\bm{1}}

\DeclareMathAlphabet{\mathsfit}{\encodingdefault}{\sfdefault}{m}{sl}
\SetMathAlphabet{\mathsfit}{bold}{\encodingdefault}{\sfdefault}{bx}{n}

\usepackage[utf8]{inputenc}
\usepackage[T1]{fontenc}
\usepackage{url}
\usepackage{amsfonts}
\usepackage{nicefrac}
\usepackage{microtype}
\usepackage{graphicx}
\usepackage{subcaption}
\usepackage{booktabs}
\usepackage{hyperref}

\usepackage{amsmath}
\usepackage{amssymb}
\usepackage{mathtools}
\usepackage{amsthm}
\usepackage{tikz}
\usetikzlibrary{calc, arrows.meta, positioning}
\usepackage{tcolorbox}
\usepackage{wrapfig}
\usepackage{tcolorbox}
\usepackage{listings}
\usepackage{xcolor}
\usepackage{xspace}
\tcbuselibrary{listingsutf8, skins, breakable}

\newtcblisting{samplebox}[2][]{
enhanced, breakable,
listing only,
listing options={
  basicstyle=\footnotesize\ttfamily,
  breaklines=true, breakatwhitespace=true,
  columns=fullflexible, keepspaces=true,
  showstringspaces=false, upquote=true,
},
colback=gray!5, colframe=gray!40, boxrule=0.4pt, arc=2pt,
left=6pt, right=6pt, top=4pt, bottom=4pt,
fonttitle=\bfseries\small, coltitle=black, colbacktitle=gray!15,
attach boxed title to top left={xshift=6pt, yshift=-7pt},
boxed title style={colframe=gray!40, colback=gray!15,
                   sharp corners, boxrule=0.3pt},
title=#2, #1
}

\newcommand{\sample}[2]{%
\begin{tcolorbox}[
  enhanced, breakable,
  colback=gray!5, colframe=gray!40, boxrule=0.4pt, arc=2pt,
  left=6pt, right=6pt, top=4pt, bottom=4pt,
  fonttitle=\bfseries\small, coltitle=black, colbacktitle=gray!15,
  attach boxed title to top left={xshift=6pt, yshift=-7pt},
  boxed title style={colframe=gray!40, colback=gray!15,
                     sharp corners, boxrule=0.3pt},
  title={#1}]
  \lstinputlisting[
    basicstyle=\footnotesize\ttfamily,
    breaklines=true, breakatwhitespace=true,
    columns=fullflexible, keepspaces=true,
    showstringspaces=false, upquote=true,
    aboveskip=0pt, belowskip=0pt,
  ]{#2}
\end{tcolorbox}
}

\newcommand\grpo{GRPO\xspace}
\newcommand\refsol{\textsc{RefSol}\xspace}
\newcommand\stepfb{\textsc{StepAlignFB}\xspace}

\definecolor{studentcolor}{HTML}{C0392B}   
\definecolor{teachercolor}{HTML}{1E8449}   
\definecolor{fbteachercolor}{HTML}{B7791F} 
\definecolor{criticcolor}{HTML}{1F4E79}    

\lstdefinestyle{promptbox}{
  basicstyle=\ttfamily\footnotesize,
  breaklines=true,
  breakatwhitespace=false,
  columns=fullflexible,
  keepspaces=true,
  showstringspaces=false,
  literate=
    {→}{$\to$}{1}
    {≡}{$\equiv$}{1}
    {≥}{$\geq$}{1}
    {≤}{$\leq$}{1}
    {…}{...}{3},
}

\newtcblisting{studentprompt}{
  enhanced, breakable,
  colback=studentcolor!5, colframe=studentcolor,
  fonttitle=\bfseries\color{white},
  coltitle=white,
  title=Solver Prompt,
  listing only,
  listing options={style=promptbox},
  boxrule=0.6pt, arc=2pt, left=4pt, right=4pt, top=4pt, bottom=4pt,
}

\newtcblisting{teacherrefprompt}{
  enhanced, breakable,
  colback=teachercolor!5, colframe=teachercolor,
  fonttitle=\bfseries\color{white},
  coltitle=white,
  title=Teacher Prompt --- \refsol (reference-solution conditioning),
  listing only,
  listing options={style=promptbox},
  boxrule=0.6pt, arc=2pt, left=4pt, right=4pt, top=4pt, bottom=4pt,
}

\newtcblisting{teacherfbprompt}{
  enhanced, breakable,
  colback=fbteachercolor!5, colframe=fbteachercolor,
  fonttitle=\bfseries\color{white},
  coltitle=white,
  title=Teacher Prompt --- \stepfb (step-aligned feedback conditioning),
  listing only,
  listing options={style=promptbox},
  boxrule=0.6pt, arc=2pt, left=4pt, right=4pt, top=4pt, bottom=4pt,
}

\newtcblisting{criticprompt}{
  listing only,
  listing utf8=latin1,
  enhanced, breakable,
  colback=criticcolor!5, colframe=criticcolor,
  fonttitle=\bfseries\color{white}, coltitle=white,
  title=Critic Prompt (Qwen/QwQ-32B),
  listing options={style=promptbox},
  boxrule=0.6pt, arc=2pt, left=4pt, right=4pt, top=4pt, bottom=4pt,
}

\begin{document}

\title{The Role of Feedback Alignment in Self-Distillation}

\author{Semih Kara$^1$ and O\u{g}uzhan Ersoy$^1$ \\
$^1$\small{Gensyn}\\}

\correspondingauthor{semih@gensyn.ai}

\begin{abstract}
Conditioning a language model on additional context, such as feedback on a previous attempt, typically improves its response. Self-distillation trains the model to retain this improvement when the context is not present. The method works by matching the model's output distribution under two settings: a \textit{student} that sees only the question, and a \textit{self-teacher} that also sees the context. What the model learns therefore depends on what context the self-teacher receives, yet the design of this context remains largely unexplored.

\vspace{0.2cm}

We study context design for self-distillation by training a solver on feedback from a frozen critic. We compare three conditions: (i) a binary reward (\grpo), (ii) the reference solution, and (iii) a step-by-step critique aligned to the solver's reasoning trace.

\vspace{0.2cm}

Step-aligned critique yields the largest gains, outperforming \grpo by 16.11 points and reference-solution-conditioned self-distillation by 5.27 points (Avg@12). Per-token advantage analysis reveals why: step-aligned feedback targets only the tokens where reasoning fails, leaving correct behavior intact. Conditioning on the reference solution, by contrast, pressures the model to change its behavior at every token (even correct steps) because an alternative derivation inevitably differs in phrasing and approach. This suggests that structural alignment between feedback and the solver's reasoning is a key driver of self-distillation effectiveness.
\end{abstract}

\maketitle

\let\thefootnote\relax\footnotetext{Accepted to the ICML 2026 Workshop on RL from World Feedback (RLxF).}

\section{Introduction}
\label{sec:intro}

Post-training with reinforcement learning has become the dominant recipe for improving LLM reasoning. The standard approach is reinforcement learning from verifiable rewards (RLVR) \citep{shao2024deepseekmath}, which learns from a single scalar reward per rollout. This reward says whether the final answer is correct, but not where in the reasoning trace the model went wrong, making credit assignment difficult. Alternatively, distillation \citep{hinton2015distilling} provides dense, token-level supervision. However, it requires access to the logits of a strong teacher, which may not exist behind an API or may be too costly to transfer at scale.

Self-distillation~\citep{hubotter2026sdpo, zhao2026opsd, shenfeld2026sdft} sidesteps both limitations. The same model plays two roles: a \textit{student} conditioned on the question alone, and a \textit{self-teacher} conditioned on the question plus some additional context $c$ (e.g., execution traces, follow-up prompts, reference solutions, feedback from another model, etc.). Training minimizes a divergence between these two distributions, distilling capabilities that emerge in-context~\citep{brown2020fewshot} into the context-free policy. 

Prior work has used various forms of context for self-distillation: code execution traces \citep{hubotter2026sdpo}, ground-truth solutions \citep{zhao2026opsd}, and feedback from other models \citep{song2026rltf} or users \citep{wang2026openclawrl,buening2026aligning}. In all cases, the context is treated as a fixed choice. No prior work, to our knowledge, studies how the design of the context affects what the model learns. Nonetheless, when the context is feedback from another model, the practitioner can design its structure. This raises the question: \textbf{what form of feedback produces the most effective self-teacher?}

We study this in the mathematical reasoning domain, using a solver–critic setup. We compare three conditions: (i) a binary reward (\grpo), (ii) the reference solution (\refsol), and (iii) a step-by-step critique aligned to the solver's reasoning trace (\stepfb). Our main findings are:
\begin{list}{\textbullet}{\leftmargin=1em \itemsep=0pt \parskip=0pt}
    \item \textbf{\stepfb outperforms \grpo and \refsol.} Step-aligned critique outperforms both \grpo and self-distillation with the reference solution on all aggregation-style accuracy metrics in our evaluation (see Section~\ref{sec:results}).
    \item \textbf{\stepfb provides implicit process supervision.} Per-token advantage analysis shows that step-aligned feedback concentrates distributional shifts at error-adjacent tokens, while solution-level feedback produces a more diffuse signal. The localization mirrors what a PRM provides~\citep{lightman2024lets, uesato2022solving}, but is obtained without training a reward model or collecting per-step scalar labels.
    \item \textbf{Feedback alignment matters as much as feedback quality.} A complete, correct reference derivation is a strong signal, but in self-distillation it diffuses across the solver's rollout because the derivation diverges from the solver's trace in surface form even at correct steps. Step-aligned critique, by addressing the solver's actual trace, concentrates the signal at the tokens where reasoning goes wrong. Self-distillation already provides a form of process-level signal through its token-level advantages. \textit{Step-aligned feedback amplifies this by concentrating the signal at the tokens where reasoning goes wrong.}
\end{list}

\section{Background}
\label{sec:background}
 
\paragraph{Distillation.}
Knowledge distillation~\citep{hinton2015distilling} trains a student model $\pi_\theta$ to match a stronger teacher $\pi_T$ by minimizing the per-token forward KL divergence:
\begin{equation}
    \mathcal{L}_{\text{KD}} = \mathbb{E}_{y~\sim~\pi_T(\cdot \mid x)}\left[\mathrm{KL}\!\big(\pi_T(y \mid x) \;\big\|\; \pi_\theta(y \mid x)\big)\right].
    \label{eq:kd}
\end{equation}
This provides dense, token-level supervision but requires access to the logits of a strong teacher model (which is often not available). Standard distillation is also off-policy: the student trains on the teacher's rollouts\footnote{In next-token prediction, a rollout is the model's complete response to a prompt.}, not its own. Thus, errors compound as the student's distribution drifts from the teacher's at inference time~\citep{ross2011reduction}. On-policy distillation~\citep{agarwal2024onpolicy,lu2025onpolicydistill} addresses this by having the student generate its own rollouts and optimizing a reverse-KL-style distillation objective using token-level supervision from the teacher, but it still requires access to the logits of a stronger teacher.
 
\paragraph{RLVR.}
Reinforcement learning with verifiable rewards removes the need for a teacher. Given a question $x$, the model samples $G$ rollouts $\{y_i\}_{i=1}^G \sim \pi_\theta(\cdot \mid x)$, each scored by a binary reward $r_i \in \{0, 1\}$. \grpo~\citep{shao2024deepseekmath} estimates per-rollout advantages via group normalization: 
\begin{align}
A_i^{\text{GRPO}} = \frac{r_i - \bar{r}}{\sigma(r)}, \label{eq:grpo}
\end{align}
where $\bar{r}$ and $\sigma(r)$ are the mean and standard deviation of rewards within the group, respectively. The policy is updated by maximizing\footnote{We present the simplified, on-policy expression (without clipping) and remove the KL penalty. We refer to \citet{shao2024deepseekmath} for the full details.}:
\begin{align*}
    &\mathcal{J}_{\text{GRPO}}(\theta) =
    \mathbb{E}_{\{y_i\}_{i=1}^{G} \sim \pi_{\theta}(\cdot \mid x)}
    \left[
    \sum_{i=1}^{G} \frac{1}{|y_i|} \sum_{t=1}^{|y_i|}
    \log \pi_\theta(y_{i,t} \mid x, y_{i,<t})\, A_{i,t}
    \right].
\end{align*}
Notice that $A_{i,t} = A_i^{\text{GRPO}}$ is constant across all tokens within a rollout.

\paragraph{Self-distillation.}
Self-distillation~\citep{hubotter2026sdpo, zhao2026opsd, shenfeld2026sdft} unifies the dense supervision of distillation with the teacher-free and on-policy properties of RL. The same model serves as both student and teacher under different prompting contexts. The \emph{student} is conditioned on the question alone, $\pi_\theta(\cdot \mid x, y_{<t})$, whereas the \emph{self-teacher} is conditioned on the question augmented with additional context $c$, giving the distribution $\pi_\theta(\cdot \mid x, c, y_{<t})$. 

Training minimizes a per-token divergence (forward KL, reverse KL, or Janson-Shannon divergence) between these two distributions, yielding the loss function:
\begin{equation}
    \mathcal{L}_{\text{SD}} = \mathbb{E}_{y~\sim~\pi_\theta(\cdot \mid x)}\left[D\!\big(\pi_\theta(y \mid x) \;\big\|\; \mathrm{sg}\left[\pi_\theta(y \mid x,c)\right] \big)\right],
    \label{eq:sd}
\end{equation}
where $D$ is a divergence and $\mathrm{sg}[\cdot]$ denotes stop-gradient. \citet{hubotter2026sdpo} show that, when $D$ is the KL divergence, the gradient of $\mathcal{L}_{\text{SD}}$ has the same form as the gradient of $-\mathcal{J}_{\mathrm{GRPO}}$, but with per-token advantages given by
\begin{equation}
    A_t^{\text{SD}}(\hat{y}_t) = \log \pi_\theta(\hat{y}_t \mid x, c, y_{<t}) - \log \pi_\theta(\hat{y}_t \mid x, y_{<t}).
    \label{eq:sd_advantage}
\end{equation}
$A_t^{\text{SD}}$ quantifies how much the context shifts the model's next-token prediction. Unlike the \grpo advantage, the self-distillation advantage varies at every token position, providing dense credit assignment automatically.
 
Self-distillation uses the \emph{same} model $\pi_\theta$ in both roles, distinguished only by the context $c$. The quality of the resulting advantage (Eq.~\ref{eq:sd_advantage}) therefore depends heavily on $c$. When $c$ is feedback from another model, the practitioner has control over its structure. The topic of this paper is the design of this structure.

\section{Methodology}
\label{sec:experiment_setup}

\subsection{Setup}
\label{sec:method}

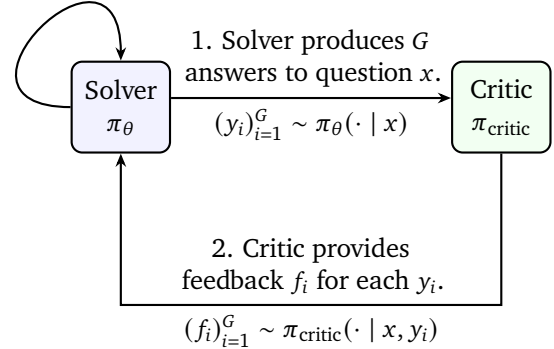
\begin{wrapfigure}{r}{0.58\textwidth}
\centering
\vspace{-0.7cm}
\hspace{-0.8cm}%
\begin{tikzpicture}[
    box/.style={draw, rounded corners=4pt, minimum height=1.2cm, minimum width=1.3cm, align=center, font=\small, thick},
    arr/.style={->, >={Stealth[length=5pt, width=4pt]}, thick},
    eq/.style={font=\small},
]
    \node[box, fill=blue!5] (solver) {Solver\\$\pi_{\theta}$};
    \node[box, fill=green!5, right=3.7cm of solver] (critic) {Critic\\$\pi_{\text{critic}}$};

    \draw[arr] ([yshift=0.12cm]solver.east) -- ([yshift=0.12cm]critic.west)
        node[midway, above, eq] {\parbox{3.5cm}{\centering 1. Solver produces $G$\\answers to question $x$.}}
        node[midway, below, eq] {$(y_i)_{i=1}^G \sim \pi_\theta(\cdot \mid x)$};

    \draw[arr] (critic.south) -- ++(0,-2) -| (solver.south);
    \path (critic.south) ++(0,-2) coordinate (corner1);
    \path (solver.south) ++(0,-2) coordinate (corner2);
    \coordinate (umid) at ($(corner1)!0.5!(corner2)$);
    \node[above, eq] at (umid) {\parbox{3.5cm}{\centering 2. Critic provides\\feedback $f_i$ for each $y_i$.}};
    \node[below, eq] at (umid) {$(f_i)_{i=1}^G \sim \pi_{\text{critic}}(\cdot \mid x, y_i)$};

    \draw[arr] (solver.west) to [out=180,in=90,looseness=5.5] (solver.north); 
        \node[above, eq] at ([xshift=0.7cm, yshift=1.4cm]solver.west) {\parbox{4cm}{\centering 3. Solver minimizes $\mathcal{L}_{\text{SD}}$\\with $(f_i)_{i=1}^G$ as context.}};
\end{tikzpicture}
\caption{Solver--critic training loop. The critic produces feedback given the question $x$ and each response $y_i$. Only the solver is trained.}
\label{fig:solver_critic}
\vspace{-0.3cm}
\end{wrapfigure}

\paragraph{Solver--critic setup.}
We implement self-distillation in a solver–critic setup (Figure~\ref{fig:solver_critic}). For each math question $x$, a trainable \emph{solver} $\pi_\theta$ generates step-tagged reasoning traces $$y = \langle\texttt{step}_1\rangle \ldots \langle\texttt{step}_N\rangle \langle\texttt{answer}\rangle.$$ A frozen \emph{critic} $\pi_{\text{critic}}$ produces feedback $f$ on the solver's response. The solver is then trained with self-distillation (Eq.~\ref{eq:sd}) using $f$ as the context $c$.
 
The only variable across our experimental conditions is $f$. The solver, loss function, divergence, and all hyperparameters remain fixed, isolating feedback structure as the independent variable.

\paragraph{Feedback conditions.}
We compare three forms of context/feedback:
\begin{list}{\textbullet}{\leftmargin=1em \itemsep=0pt \parskip=0pt}
    \item \textbf{\grpo (no feedback.)} Standard RLVR baseline. The solver generates $G$ rollouts per question, each scored with a binary reward. Advantages are group-normalized rewards (Eq.~\ref{eq:grpo}). No critic, no self-distillation.
    \item \textbf{\refsol (reference solution).} A chain-of-thought reference solution produced by a stronger model, either generated on-the-fly by the critic, or precomputed during dataset construction as by \citet{li2024numina, moshkov2025openmathreasoning}. \refsol mirrors the setup of \citet{zhao2026opsd}.
    \item \textbf{\stepfb (step-aligned critique).} The critic receives both the solver's step-tagged response and the dataset's reference solution, then produces per-step feedback. We prompt it to copy correct steps verbatim and fix incorrect or incomplete steps while staying as close to the solver's reasoning trace as possible. We observed that copying correct steps verbatim activates the model's in-context copying behavior \citep{olsson2022incontextlearninginductionheads}, which sharpens advantage estimates, especially for correct steps. We give more details in Section~\ref{sec:credit_assignment}.
\end{list}

\subsection{Training Details}
\label{sec:training_details}

Our training recipe follows \citet{zhao2026opsd} with three modifications: no per-token KL clipping, increased student rollout cap, and a critic-conditioned teacher in the \stepfb experiments.

\paragraph{Self-distillation recipe.}
We use Qwen3-1.7B \citep{yang2025qwen3technicalreport} as the solver. All rollouts are sampled on-policy, at temperature $T{=}1.1$ with a maximum token cap of $2048$. We use a group size of $G=1$ for self-distillation experiments, and set $G=8$ for \grpo). Divergence measure is forward KL. We use the Thinking-Mode-Off student / Thinking-Mode-On teacher pairing identified as optimal by \citeauthor{zhao2026opsd}. The teacher is held fixed at the initial (LoRA-free) base policy by disabling adapters at the teacher forward; no separate teacher checkpoint is loaded. LoRA \citep{hu2021loralowrankadaptationlarge} adapters are trained on all attention and MLP projections ($r{=}64$, $\alpha{=}128$) with AdamW \citep{loshchilov2019decoupledweightdecayregularization}, learning rate $5\!\times\!10^{-6}$, effective batch size $32$, bf16 precision, gradient checkpointing, and Flash Attention 2 \citep{dao2022flashattentionfastmemoryefficientexact}. Distributed training uses DeepSpeed ZeRO-2 with CPU optimizer offload \citep{rasley2020deepspeed} across 4 H100 GPUs; on-policy generation is served in-process by vLLM \citep{kwon2023efficientmemorymanagementlarge} in colocate mode. Details of the training configuration are in Appendix~\ref{app:training_details}.

\paragraph{Step-aligned feedback variant (\stepfb).}
In the step-aligned feedback variant, the teacher conditions on the critique $f$ (for a group size of 1, we denote $f:=f_1$) instead of the raw reference solution. The critique is produced once per rollout by a frozen reasoning model (Qwen/QwQ-32B; greedy decoding, $T{=}0$, Thinking-Mode-On) served via vLLM. We send a single user-message prompt that contains schema-driven grading instructions, the problem, the reference solution, and the student rollout; we strip the reasoning trace (\texttt{<think>...</think>}) from the returned content and splice only the structured grader output into the teacher context. The detailed critic prompt template is in Appendix~\ref{app:prompt_templates}. The critic prompt is long, by design, to keep its feedback actionable and aligned step-by-step with the solution. Nonetheless, vLLM's automatic prefix caching substantially mitigates the resulting overhead.

\paragraph{Dataset.}
We train on a difficulty and formatting-filtered subset of OpenMathReasoning \citep{moshkov2025openmathreasoning}. Prior work suggests that self-distillation's advantage over RLVR methods is most pronounced on harder problems. Since our focus is on how the \emph{form} of feedback affects self-distillation, we deliberately target this harder-question regime, where self-distillation is the more relevant choice over RLVR. We additionally require that the problems be tractable for the trained Qwen3-1.7B, so that the model receives \emph{actionable feedback} rather than failing outright and being handed a full reference-like solution by the critic (which would have collapsed \refsol and \stepfb into the same setup). We also require that the critic itself can solve the problem, ensuring that its feedback is coherent and high quality. Concretely, we start from the subset of OpenMathReasoning with \texttt{pass\_rate\_72b\_tir} > 0.85, then compute Avg@16 and average formatting accuracy over 16 attempts from the 1.7B model, keeping only problems where Avg@16 < 5/16 and the formatting accuracy is greater than 0.9. This yields 312 samples, of which we reserve 30 for evaluation and use the remaining 282 for training. We train for up to 7 epochs, saving checkpoints every 10 optimizer steps.

\section{Results}
\label{sec:results}

\subsection{Accuracy Results}

We evaluate the three feedback variants on a held-out 30-sample test split from OpenMathReasoning. We report aggregation-style accuracy metrics over $n=12$ samples per problem, averaged across the test set: Avg@12 (mean fraction of correct samples), Majority-Vote@12 (correctness of the majority answer), and Pass@12 (fraction of problems with at least one correct sample), alongside mean answer length.

We compare \grpo against the self-distillation variants (\refsol and \stepfb) at matched generation compute. \grpo requires $G{=}8$ rollouts per prompt for its group-relative advantage estimate, while self-distillation uses a single rollout ($G{=}1$); at equal compute, thus \grpo consumes $8\times$ fewer prompts per step. Nonetheless, each method trains for 7 epochs, ruling out the reduced dataset exposure as a confound. 

\begin{figure}[t]
\centering
\begin{subfigure}[t]{0.48\linewidth}
  \includegraphics[width=\linewidth]{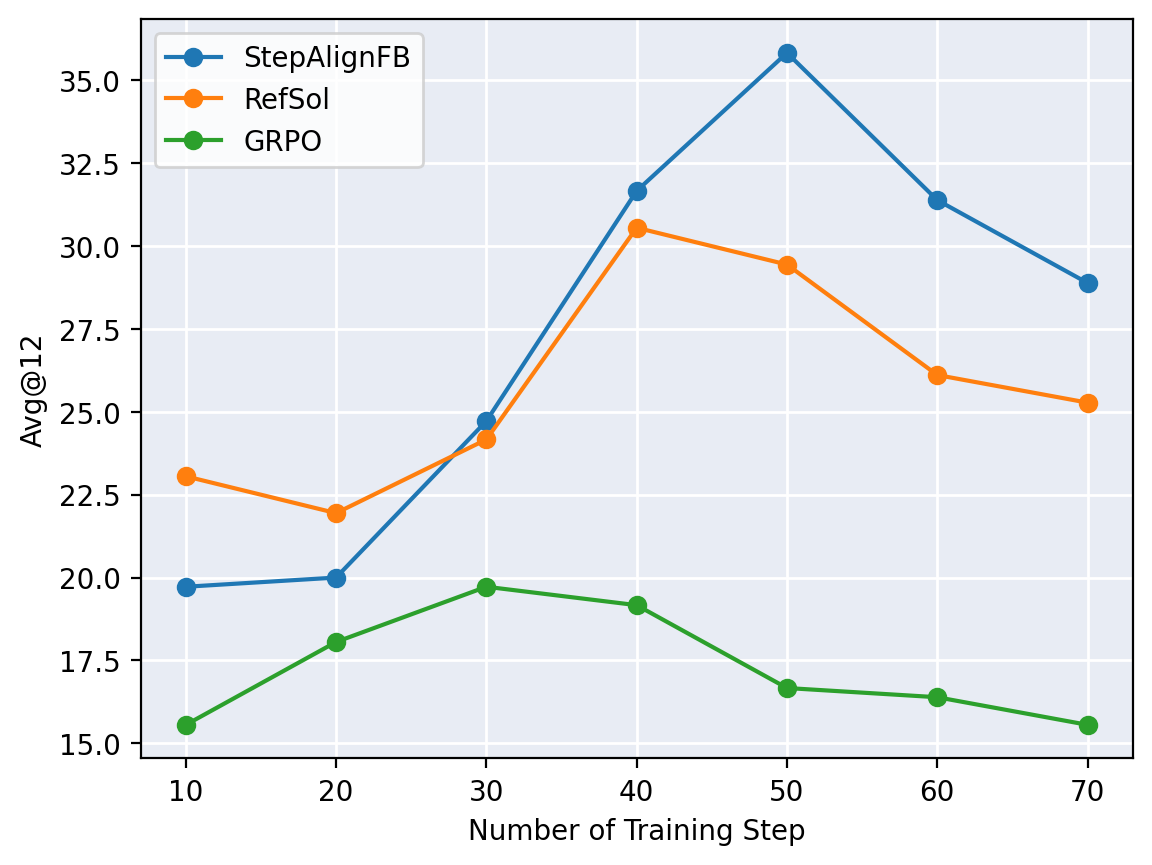}
  \caption{Avg@12}
  \label{fig:avg_at_n}
\end{subfigure}\hfill
\begin{subfigure}[t]{0.48\linewidth}
  \includegraphics[width=\linewidth]{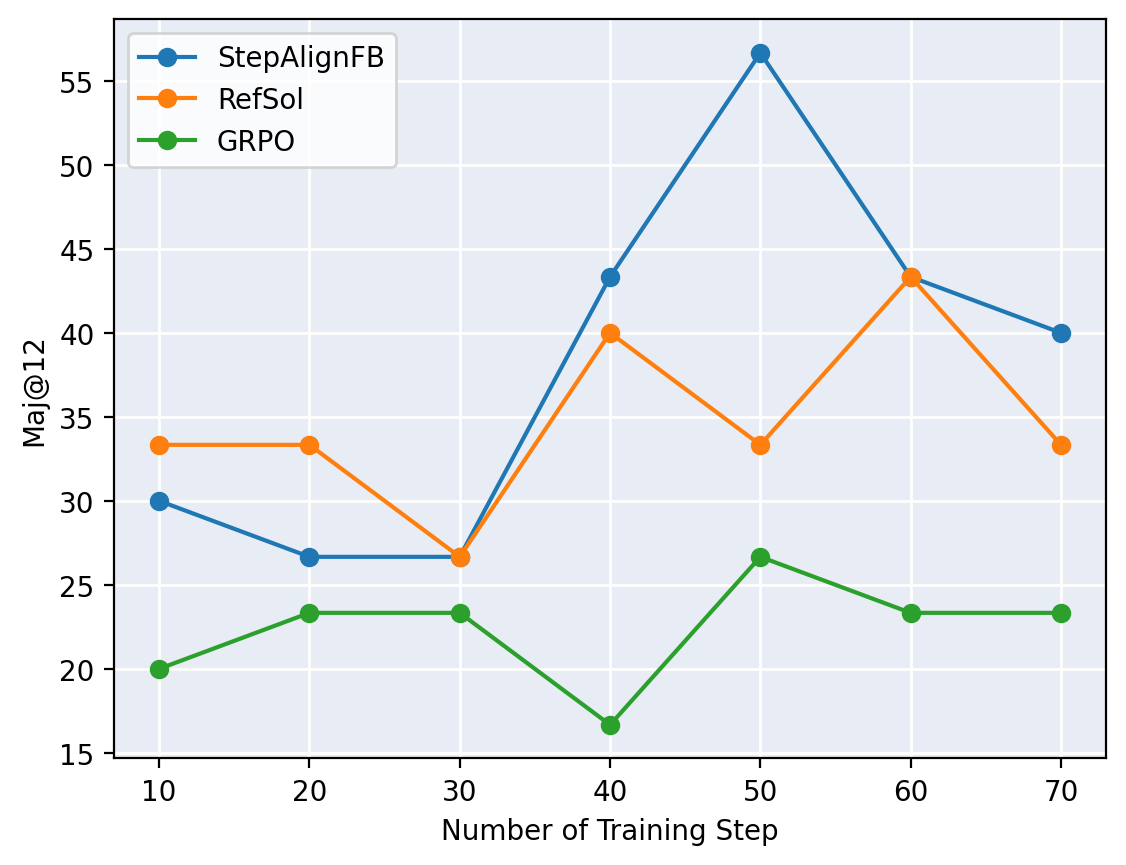}
  \caption{Majority-Vote@12}
  \label{fig:maj_at_n}
\end{subfigure}

\vspace{0.8em}

\begin{subfigure}[t]{0.48\linewidth}
  \includegraphics[width=\linewidth]{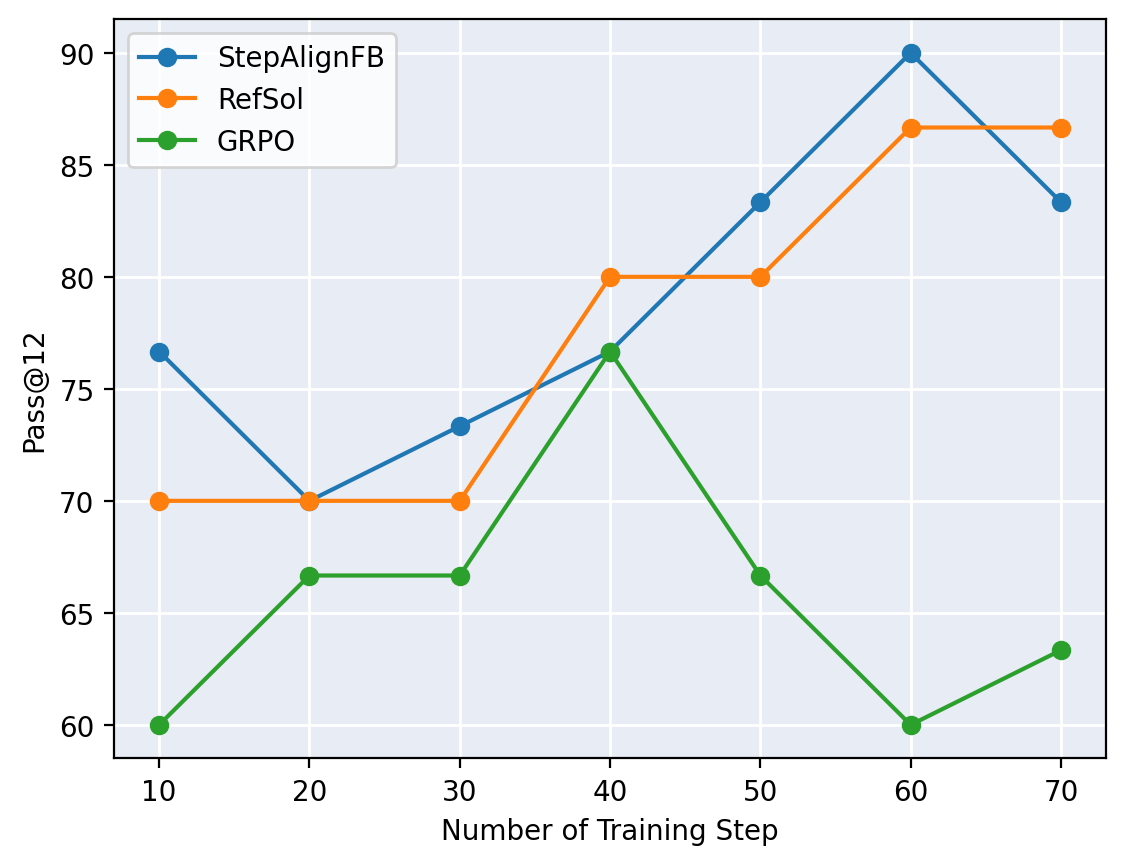}
  \caption{Pass@12}
  \label{fig:pass_at_n}
\end{subfigure}\hfill
\begin{subfigure}[t]{0.5\linewidth}
  \includegraphics[width=\linewidth]{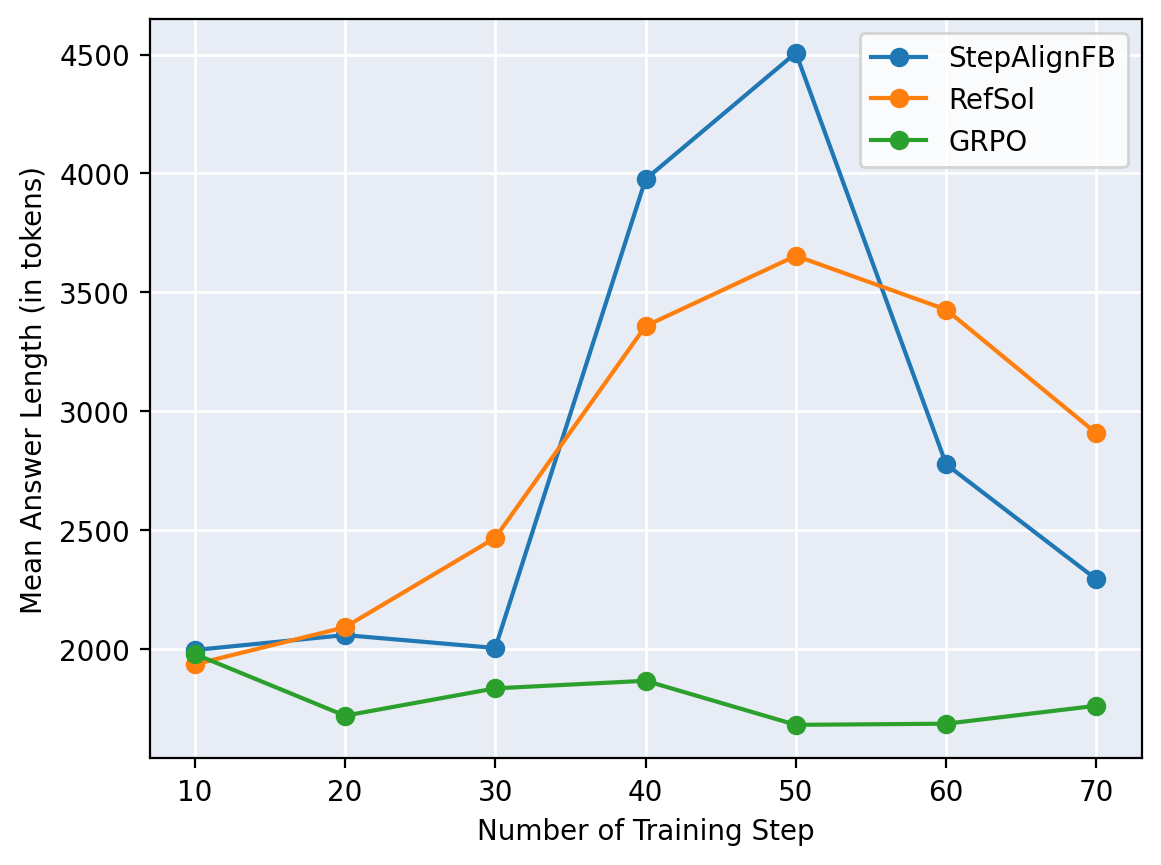}
  \caption{Mean answer length}
  \label{fig:answer_len}
\end{subfigure}
\caption{Self-distillation matches or exceeds \grpo across training on OpenMathReasoning. \refsol conditions the teacher on the reference solution provided in the OpenMathReasoning dataset; \stepfb substitutes step-level feedback; \grpo is the reward-only baseline.}
\label{fig:openmath_curves}
\end{figure}

Figure~\ref{fig:openmath_curves} plots each metric as a function of the training step. Table~\ref{tab:perf} summarizes the \textit{per-metric best value attained by each method across checkpoints} (steps 10--70), with the step at which the maximum (minimum for answer length) occurred annotated as \textit{s}. Because the methods reach peak evaluation performance at different steps, we select the best checkpoint independently for each metric rather than reporting end-of-run values or fixing a single checkpoint per method.

\begin{table}
\centering
\caption{\textbf{Per-metric best evaluation results on OpenMathReasoning.} For each (method, metric) pair we report the best value across all training checkpoints and annotate the step (\textit{s}) at which the maximum (or minimum, for answer length) was attained. Bold marks the per-column winner.}
\label{tab:perf}
\begin{tabular}{lllll}
\toprule
 & Pass@12 & Maj@12 & Avg@12 & Mean Answer Length \\
\midrule
\grpo & 76.67 (s=40) & 26.67 (s=50) & 19.72 (s=30) & \textbf{1681.49 (s=50)} \\
\refsol & 86.67 (s=60) & 43.33 (s=60) & 30.56 (s=40) & 1935.83 (s=10) \\
\stepfb & \textbf{90.00 (s=60)} & \textbf{56.67 (s=50)} & \textbf{35.83 (s=50)} & 1996.07 (s=10) \\
\bottomrule
\end{tabular}
\end{table}

\paragraph{Observations.}
We see that using step-aligned feedback as context outperforms feeding the reference solution. Despite never seeing the ground-truth derivation, \stepfb outperforms \refsol on the Avg@12 metric at the per-metric best values: +2.33 Pass@12, +13.33 Maj@12, and +5.27 Avg@12 (Table~\ref{tab:perf}). The large Majority-Vote gain in particular suggests \stepfb's policy concentrates probability on correct answers more sharply, rather than merely covering them, which is the regime that benefits most from test-time aggregation.

As with other post-training methods \citep{stollenwerk2022adaptivefinetuningtransformerbasedlanguage,zhao2026opsd}, self-distillation benefits from early stopping. We trained the model for 7 epochs but found that 5–6 sufficed to reach peak performance across all metrics. This is promising for compute efficiency, however carries a methodological caveat: a fixed end-of-run evaluation can systematically understate self-distillation's ceiling, so per-checkpoint selection on a held-out validation set is necessary for a fair comparison.

Finally, both distillation methods dominate \grpo throughout training at all fronts except answer length. While \grpo maintains token efficiency, both \refsol and \stepfb stay above \grpo in terms of accuracy throughout training, with a final Avg@12 gap on the order of 8 points.

\subsection{Token-Level Credit Assignment}
\label{sec:credit_assignment}

In this section, we identify the mechanism by which \stepfb outperforms \refsol. The self-distillation advantages (Eq.~\ref{eq:sd_advantage}) tell a clear story: step-aligned feedback causes the self-distillation signal to behave like a process reward model, reinforcing correct steps in the solver's trace while suppressing erroneous ones. Feeding the reference solution as feedback, by contrast, suppresses even fully correct solver trajectories. We present sample problems, student and reference solutions, critiques, and the resulting per-token advantage plots in Appendix~\ref{app:samples}.

\begin{figure*}[h!]
\centering
\includegraphics[width=\linewidth]{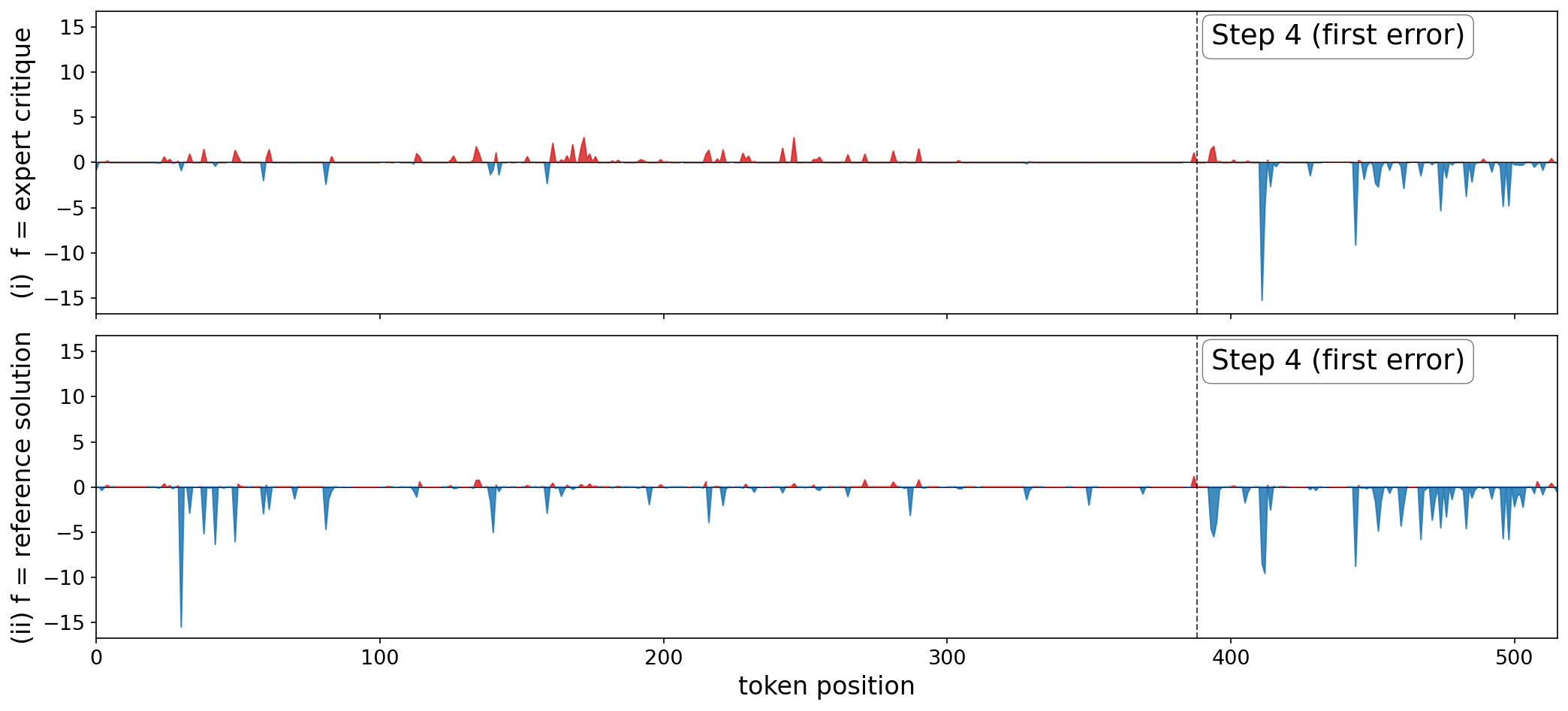}
\caption{\textbf{Incorrect step.} Per-token advantages for a rollout containing an error at the step marked by the dashed line. \stepfb (top) produces a sharp negative shift at the tokens after the erroneous step, while reinforcing correct reasoning in the preceding steps. \refsol (bottom) produces broadly negative advantages across the entire trace, conflating the error with everywhere the student's path diverges from the canonical solution.}
\label{fig:adv-error}
\end{figure*}

\textbf{\stepfb concentrates signal at errors and reinforces correct steps.} At incorrect steps, the self-teacher diverges sharply from the student, producing large negative advantages (Figure~\ref{fig:adv-error}). At correct steps, including those preceding the error, the critic faithfully preserves the solver's reasoning path. The result is a targeted learning signal that penalizes errors while reinforces the valid trace that surrounds them. This is analogous to a process reward model, but derived entirely from natural-language feedback, with no reward model training required.

\begin{figure*}[h!]
\centering
\includegraphics[width=\linewidth]{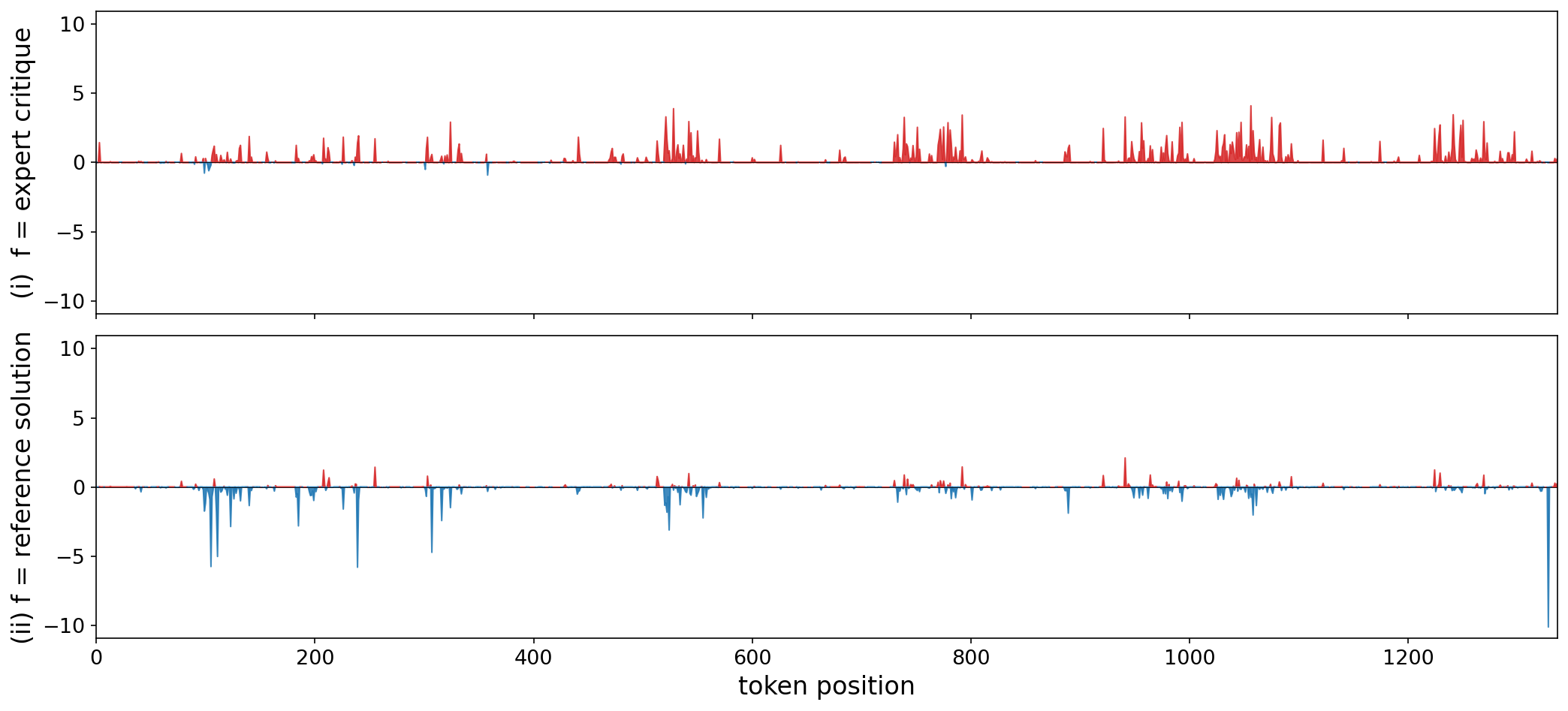}
\caption{\textbf{Fully-correct rollout.} Per-token advantages $A_t^{\text{SD}}(\hat{y}_t)$ (see Eq.~\ref{eq:sd_advantage}) along a single student trajectory that reaches the correct boxed answer. \stepfb (top) produces positive advantages throughout the trace, reinforcing the student's correct reasoning trace. \refsol (bottom) produces diffuse negative advantages across the entire rollout: the teacher prefers the alternative derivation's surface form even though the student's answer is correct, mixing stylistic disagreement into the gradient.}
\label{fig:adv-correct}
\end{figure*}

\textbf{Solution-level feedback produces diffuse suppressive signal.} Under \refsol, the self-teacher sees a complete alternative derivation: one that reaches the right answer but follows a different reasoning path, uses different notation, and phrases steps differently. Even at steps where the solver was correct, the alternative solution diverges in surface form, creating diffuse negative advantages at almost all tokens (see Figures~\ref{fig:adv-correct} and~\ref{fig:adv-error}). The self-teacher thus mixes genuine error correction with stylistic disagreement, diluting the learning signal.

\textbf{Connection to process vs.\ outcome supervision.} \citet{lightman2024lets} and \citet{uesato2022solving} show that per-step scalar rewards outperform per-problem rewards for math reasoning. Our finding is a distributional generalization: step-aligned textual feedback provides richer-than-scalar process supervision, and the self-distillation mechanism converts it into per-token credit, all without training a reward model. 

\begin{figure*}[h!]
\centering
\includegraphics[width=\linewidth]{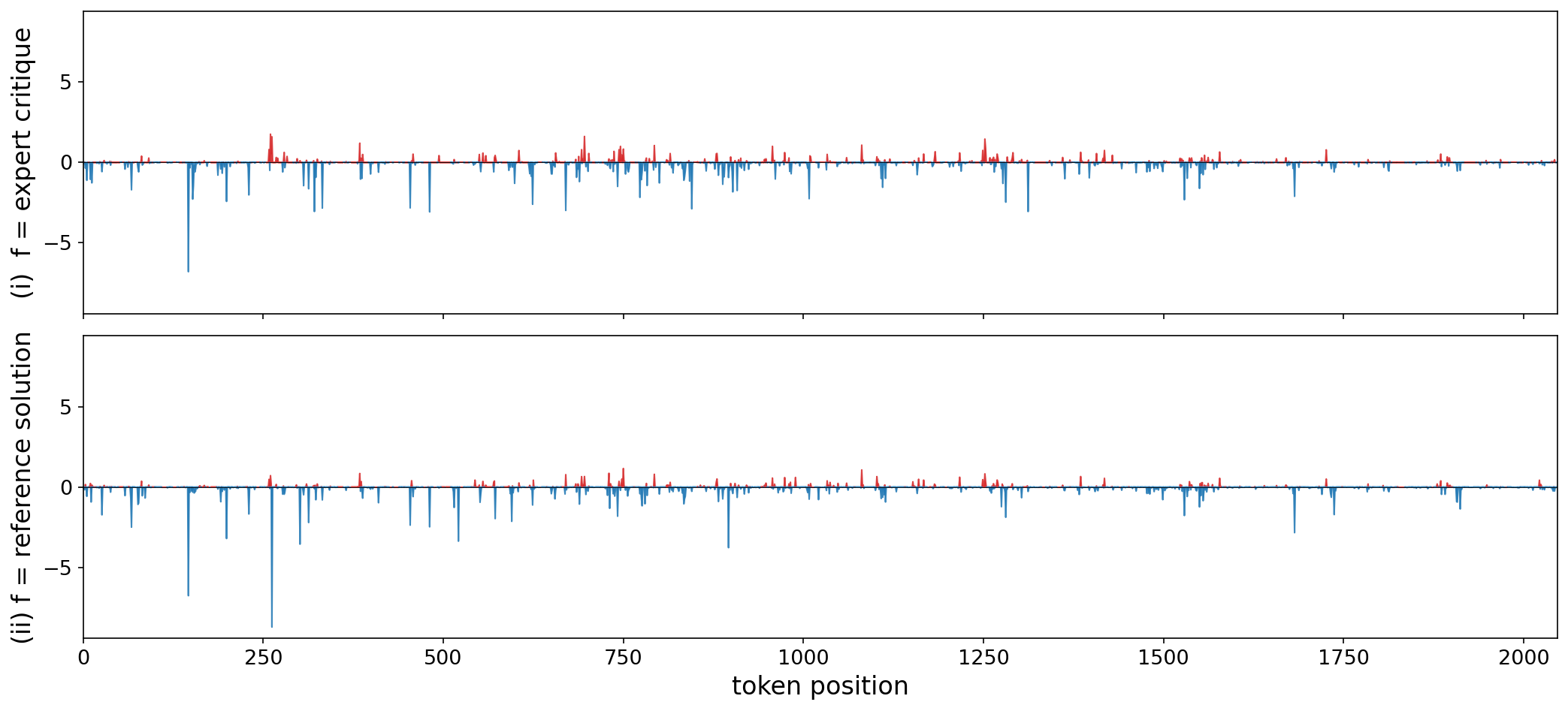}
\caption{\textbf{Incomplete solution.} Per-token advantages for a rollout that exhausted the token budget without producing a final answer. The student's reasoning trace is unpromising from the start and quickly devolves into guessing values for the variable before running out of budget. The critic responds by producing a complete alternative solution, yielding update signals similar to those of the reference solution.}
\label{fig:adv-incomplete}
\end{figure*}

\subsection{Verbatim Repetition and the Role of Induction-Head Copying}
\label{sec:induction_heads}

Our experiments showed that PRM-like credit assignment requires careful design of what the critic includes in its feedback. How much of the solver's output the critic repeats verbatim determines the quality of the advantage signal. We identify three regimes in the advantage plots. These led us to the ``faithful-scribe convention'': the critic repeats correct steps verbatim and rewrites only the incorrect ones (Hard Rules~3 and~4 in Appendix~\ref{app:prompt_templates}).

\begin{figure*}[ht]
\centering
\includegraphics[width=\linewidth]{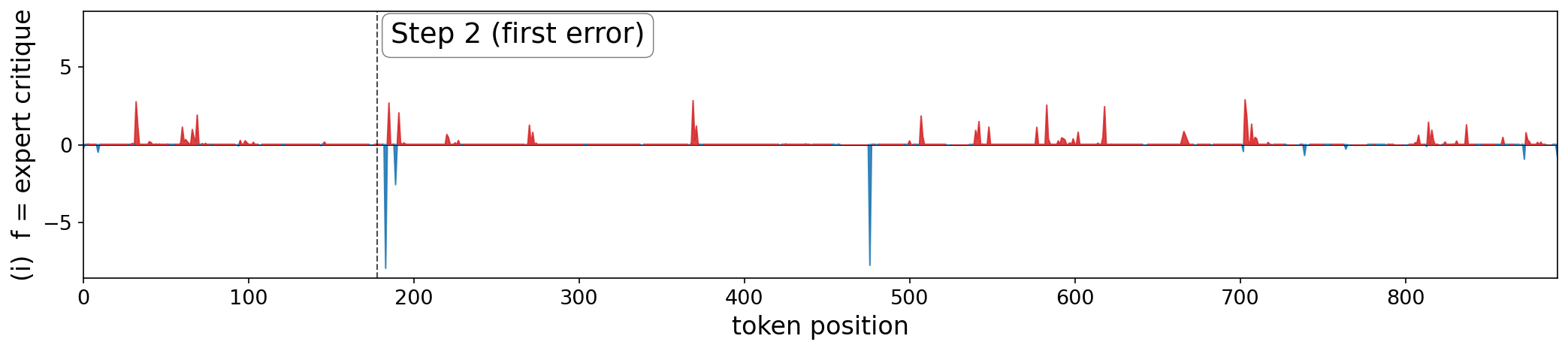}
\caption{\textbf{Full verbatim repetition with appended correction.} The critic includes the student's incorrect step verbatim and follows it with a textual correction. Induction-head copying of the original (incorrect) tokens dominates the
teacher's distribution; per-token advantages are overwhelmingly positive.}
\label{fig:adv-verbatim-error}
\end{figure*}

\paragraph{Full verbatim repetition overrides the corrective signal.} A natural first question is whether to expose the full original attempt in the feedback: \emph{``Here is your previous attempt: \texttt{<previous\_attempt>}. Step $N$ is incorrect; the correct step should be \texttt{<corrected\_step>}.''} Consistent with \citet{hubotter2026sdpo}, we found this consistently fails. The verbatim quote, placed earlier in the context, anchors the model at the erroneous step: the teacher reproduces tokens from the original (incorrect) attempt, and the per-token advantages are diffusely positive (Figure~\ref{fig:adv-verbatim-error}). The corrective text appended after the quote does not override the in-context copy, and the gradient do not see the error.

\begin{figure*}[ht]
\centering
\includegraphics[width=\linewidth]{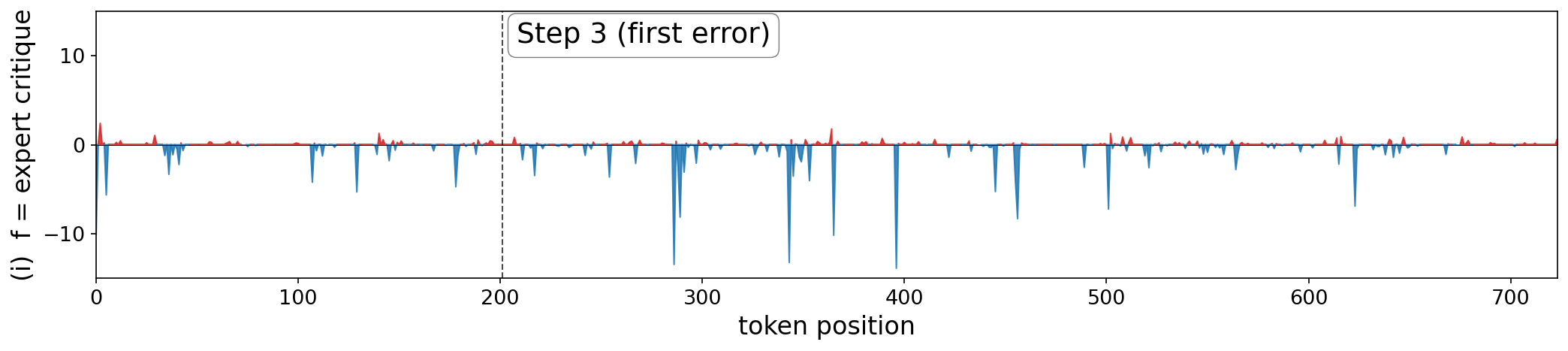}
\caption{\textbf{Solution omitted from the teacher context.} For a fully correct student solution, the critic acknowledges correctness without reproducing the trace. With no in-context anchor on the student's tokens, the teacher's distribution
drifts toward alternative surface forms, producing negative advantages across the correct trajectory.}
\label{fig:adv-no-repeat}
\end{figure*}

\paragraph{Omitting the solution suppresses correct reasoning.} The dual failure mode arises when we don't provide any of the correct steps from the original attempt (verbatim). A natural shorthand is to acknowledge correctness without reproducing the trace, e.g. \emph{`This step is correct.''} However, without the student's tokens in the context, the self-teacher distribution can drift away from the student's at the positions the student got right. The per-token advantages on the correct trace come out diffusely negative (Figure~\ref{fig:adv-no-repeat}), echoing the diffuse-suppression pattern \refsol produces in Figure~\ref{fig:adv-correct} despite the feedback endorsing the student.

\paragraph{Partial verbatim repetition up to the first error preserves both properties.} The configuration that works is to repeat the student's solution verbatim only up to (but not including) the erroneous step, replace that step with the correct claim, and continue from there. The teacher's distribution then aligns with the student's tokens on the correct prefix (producing positive advantages), while the absence of an in-context anchor at the erroneous position lets the corrected continuation govern the teacher's predictions there (producing the sharp negative cluster). This is the pattern shown in Figure~\ref{fig:adv-error}, and it is precisely the behavior our critic prompt is engineered to elicit.

\paragraph{A mechanistic hypothesis: induction-head copying.} These three regimes mirror the prefix-matching-and-copying behavior of induction heads characterized by \citet{olsson2022incontextlearninginductionheads}. An induction head implements
the pattern $[A][B]\ldots[A]\to[B]$: when a token sequence appears earlier in the context, attending back to it increases the logit of the token that followed it, \textit{biasing the model toward repeating the in-context continuation}. Under this
account, a full verbatim quote of the student's incorrect step installs the erroneous continuation as the in-context $[B]$; subsequent corrective text arrives too late and too disconnected from the matching prefix to override the copy. Partial repetition is the only configuration that uses copying \emph{selectively}: the correct prefix is anchored, so induction-head
copying reinforces it; the erroneous step is left un-anchored, so the corrected continuation freshly written by the critic governs the teacher's predictions there. In effect, the faithful-scribe convention recruits induction-head copying as a
tool for credit assignment rather than a source of error reinforcement.

\citet{hubotter2026sdpo} report a closely related effect in a different setting. Ablating whether the student's original attempt $y$ is included in the teacher template, they find that ``including it biases the teacher towards the student's attempt \ldots\ thereby reducing exploration'' (their Table~6), and that ``naively including only solutions or initial attempts $y$ significantly reduces diversity in the teacher and student.'' Our findings are consistent with this result and we hypothesize that in-context copying via induction heads is the underlying mechanism. The practical implication, however, is sharper than ``do not repeat the attempt.'' Repetition is not uniformly harmful; it can be harmful only where the student's tokens are wrong. Repeating the \emph{correct prefix} while withholding the erroneous step converts the copying bias from a liability into the mechanism that concentrates the self-distillation signal at errors. This recovers, by prompt design alone, the localization property that PRMs obtain through step-labeled training.

\section{Related Work}

\textbf{Self-distillation for LLMs.} Main references for self-distillation are \citet{hubotter2026sdpo,shenfeld2026sdft,song2026rltf,zhao2026opsd}. \citet{hubotter2026sdpo} introduced SDPO and used code execution trace, reference solutions from datasets, and stronger models' solutions as feedback. OPSD \citet{zhao2026opsd} used ground-truth solutions as teacher context. RLTF \citep{song2026rltf} formalized learning from text feedback using both self-distillation and feedback-modeling objectives, though they re-sample in-context rollouts from the solver. SDFT \citep{shenfeld2026sdft} applied self-distillation to continual learning from demonstrations.

\textbf{Process vs. outcome supervision.} PRMs \citep{lightman2024lets} and process-based feedback \citep{uesato2022solving} assign per-step scalar rewards and consistently outperform outcome supervision \citep{cobbe2021trainingverifierssolvemath} for math. The advantage is twofold: step-level signals localize errors to the exact reasoning step where they occur, which provides denser credit assignment than a single end-of-trajectory reward, and they discourage models from reaching correct answers via flawed intermediate steps. To avoid the cost of human step-level annotation, follow-up work has automated the labeling process by estimating each step's value from rollouts to the final answer \citep{wang2024mathshepherd}. PRMs are widely used both as verifiers for best-of-N selection and as dense reward signals for RL.

\paragraph{Multi-agent debate and LLM-as-judge.} Multi-agent debate~\citep{du2023debate} and LLM-as-judge~\citep{zheng2023judging} introduced inter-model interaction for improving reasoning at inference time. Since then, LLM-generated judgments have been widely adopted as training signals~\citep{yuan2024selfrewarding, chan2023chateval, zhao2025absolutezero}. Our solver--critic setup follows this paradigm.

\textbf{Structured reasoning.} Chain-of-thought prompting \citet{wei2022cot}, along with the closely related scratchpad approach \citep{nye2021scratchpad}, established that exposing intermediate computations dramatically improves multi-step reasoning, and that prompting alone is often enough to elicit it \citep{kojima2022zeroshot}. STaR \citep{zelikman2022star} extends this by turning CoT from a prompting trick into a training signal: it samples rationales from the model, keeps those that yield the correct answer, and fine-tunes on them, iterating to bootstrap stronger reasoning from the model's own outputs.

\section{Conclusion}

As demonstrated with experiments, the structure of feedback is a central determinant of self-distillation effectiveness. Among the three training signals we study, step-aligned feedback (\stepfb) consistently performs best on all accuracy metrics, outperforming both the sparse binary reward (used in \grpo) and self-distillation with reference solution (\refsol). This comparison highlights a clear trade-off. Firstly, as observed by prior work as well \citep{hubotter2026sdpo,zhao2026opsd,song2026rltf}, \grpo is simple and inexpensive, but its scalar reward provides only weak credit assignment over long reasoning traces. Secondly, \refsol offers richer supervision than \grpo, yet because it is not explicitly aligned to the solver’s intermediate steps, its learning signal remains diffuse. By contrast, \stepfb preserves the token-level benefits of self-distillation while adding structural alignment to the solver’s reasoning process, yielding an implicit form of process supervision that concentrates around error-prone parts of the trace. At the same time, this stronger signal comes with an important practical drawback: producing high-quality step-aligned critiques requires a capable critique model, which can substantially increase training cost and system complexity relative to the other methods. Reference solutions with chain-of-thought are also commonly sampled from strong models. Nonetheless, unlike solution-specific feedback, they are not tailored to an individual rollout and can be reused across many training runs.

Our findings suggest that the main advantage of self-distillation is not merely access to additional context, but also access to context whose structure matches the reasoning trajectory the student. Our study, however, is limited in scope: all experiments are conducted on the OpenMathReasoning dataset, so it remains unclear how well these conclusions transfer to other mathematical benchmarks, broader reasoning domains, or different solver--critic configurations. Future work should therefore test whether the benefits of step-aligned critique persist across datasets, model sizes, and tasks, and whether similarly effective but cheaper critique-generation strategies can narrow the cost gap that currently limits the practicality of \stepfb.

\bibliographystyle{plainnat}
\bibliography{references/semih_references}


\appendix

\section{Experiment details}

\subsection{Prompt templates}
\label{app:prompt_templates}

\begin{figure}[ht]
\centering
\begin{studentprompt}
Problem:
{problem}

Solve the problem step by step. Format each step as:

### Step N: <topic>
<reasoning for step N>

Please reason step by step, and put your final answer within \boxed{}.
\end{studentprompt}
\caption{\textbf{Solver prompt.} The solver sees only the problem statement, with the same step-formatted instruction used by the evaluation harness so that on-policy training and inference share a prompt template.}
\label{fig:student-prompt}
\end{figure}

\begin{figure}[ht]
\centering
\begin{teacherrefprompt}
Question: {problem}

A reference solution is given below. Use it only to ensure correctness.

Reference solution:
{reference_solution}

Instructions:
- Produce a fresh, self-contained solution to the original problem.
- Use the reference solution only to ensure correctness; do not mention or refer to it.

Let's think step by step and produce a final answer in the format \boxed{}.
\end{teacherrefprompt}
\caption{\textbf{Teacher prompt for \refsol.} The reference solution provided in the dataset is the privileged context.}
\label{fig:teacher-prompt-ref}
\end{figure}

\begin{figure}[ht]
\centering
\begin{teacherfbprompt}
Question: {problem}

Expert feedback on a prior attempt at this problem is given below. The feedback diagnoses where the attempt went wrong (by step number) and carries the corrected continuation.

Expert feedback:
{expert_critique}

Instructions:
- Produce a fresh, self-contained solution to the original problem.
- Use the feedback only to ensure correctness; do not mention or refer to it.

Let's think step by step and produce a final answer in the format \boxed{}.
\end{teacherfbprompt}
\caption{\textbf{Teacher prompt for \stepfb.} The reference solution is replaced by a critic-generated critique $f$ produced by Qwen/QwQ-32B (Figure~\ref{fig:critic-prompt}).}
\label{fig:teacher-prompt-fb}
\end{figure}

\clearpage
\begingroup
\par\medskip
\begin{criticprompt}
You are a math grader producing feedback to a student's solution.
    Your default behavior is faithful scribe: when the student's work is correct up to some point, reproduce that portion. The exceptions:
    - In Case D, at the erroneous step only, replace the student's claim with the correct one.
    - In Case C, you may compress the student's correct steps because the student ran out of room -- this exception applies ONLY to Case C and nowhere else.

    # Decision procedure

    **Step 1: Classify into ONE case.** Check in order, stop at first match:
    - No final answer (no \\boxed{{}}, trails off, stops mid-derivation) -> Case C
    - Boxed answer doesn't match reference (allow equivalent forms: 1/2 == 0.5) -> Case D
    - Boxed answer matches but a non-routine (non-routine = named theorem, substitution, key equation, or non-algebraic claim) step is unjustified or invalid -> Case B
    - Else -> Case A

    **Step 2: Identify the pivotal step N.**
    - B: step with the unjustified claim.
    - C: last step before stopping.
    - D: earliest erroneous step (not inherited errors).

    **Step 3: Output the matching schema below. No preamble, no postamble, no commentary outside the schema.**

    # Schemas

    Every schema begins with a `### Summary` block giving a one-line verdict per student step. The body that follows uses headers and `---` delimiters to make the structure of the feedback explicit.

    ## Case A

    ### Summary
    Step 1: Correct.
    Step 2: Correct.
    ...
    Step <final>: Correct.

    # Your solution is fully correct.
    ---
    <Reproduce the student's solution, including the \\boxed{{}} answer.>
    ---

    ## Case B

    ### Summary
    Step 1: Correct.
    Step 2: Correct.
    ...
    Step <N>: Correct, but missing justification -- <one phrase naming what's missing>.
    Step <N+1>: Correct.
    ...
    Step <final>: Correct.

    # Your solution reaches the correct answer, but Step <N> is missing justification.
    # This is the part of your solution before the gap:
    ---
    <Student's steps 1 through N-1, reproduced.>
    ---
    # Step <N> needs the following justification added:
    ---
    <Student's step N copied, with the missing justification -- named theorem, formula, or equation -- appended.>
    ---
    # The remainder of your solution is correct:
    ---
    <Student's steps N+1 through final, reproduced, including the \\boxed{{}} answer.>
    ---

    ## Case C

    ### Summary
    Step 1: Correct.
    Step 2: Correct.
    ...
    Step <N>: Correct, but stopped here without producing a final answer.

    # Your solution was correct up to Step <N> but ran out of room before finishing. Below is a condensed version of your work followed by the completion.
    # This is the correct part of your solution, condensed:
    ---
    <Student's steps 1 through N, condensed.>
    ---
    # Here is the rest of the solution:
    ---
    <Continue the derivation in the student's notation, concisely. One operation per step.>
    ---
    # Final answer
    \\boxed{{<answer>}}

    ## Case D

    ### Summary
    Step 1: Correct.
    Step 2: Correct.
    ...
    Step <N>: Incorrect -- <one phrase naming the error>.
    Step <N+1>: <verdict>.
    ...
    Step <final>: <verdict>.

    # Your solution has an error at Step <N>. Below is your correct work, the corrected step, and the remainder of the solution.
    # This is the correct part of your solution:
    ---
    <Student's steps 1 through N-1, reproduced.>
    ---
    # You made an error at Step <N>. Here is the corrected step:
    ---
    Step <N>: <Write the correct claim in the student's style -- same notation, same step granularity, same level of detail as the surrounding correct steps. Do NOT reproduce the student's incorrect step.>
    ---
    # Continuing the solution from the corrected step:
    ---
    <Continue the derivation correctly from the corrected step N, in the student's notation and style. Include the final \\boxed{{}} answer.>
    ---

    # Hard rules

    1. Output ONE schema only. Every schema starts with `### Summary`. No preamble before `### Summary`, no postamble after the final delimiter.
    2. The `### Summary` block contains one line per student step in the format `Step <i>: <verdict>` where verdict is "Correct" or "Incorrect -- <one phrase>" or a similar terse marker.
    3. In Cases A, B, and D, your reproduction of the student's correct steps should follow same or similar equation, notations, wording. If the student's reasoning trace correct, do not deviate from the student's solution.
    4. In Case D, at step N only, replace the student's incorrect claim with the correct claim in the student's style. Correct steps before and after are copied exactly per rule 3.
    5. Every hypothesis in the problem must appear in the steps at a named step.
    6. No advisory phrases ("be more rigorous", "needs justification") unless immediately followed by the specific content.
    7. Equivalent approaches reaching the same answer are CORRECT. The reference is an answer key, not a required path.
    8. In Case D, continuations after step N match the student's style: same notation, same step granularity, same level of formality. In Case C, continuations are written concisely -- one operation per step, no exploration.
    9. Don't invent formulas absent from the reference or not derivable from the problem.

    Problem:
    {problem}

    Reference answer:
    {reference_solution}

    Student's solution:
    {student_sol}
\end{criticprompt}
\captionof{figure}{\textbf{Critic prompt.} A single user-turn message sent to a frozen Qwen/QwQ-32B. The grader is instructed to classify the student rollout into one of four cases (A: fully correct; B: correct answer with an unjustified non-routine step; C: ran out of tokens; D: incorrect answer) and emit feedback in a schema specific to that case. The returned reasoning trace (\texttt{<think>...</think>}) is stripped before the structured grader output is spliced into the final feedback $f$ for the \stepfb teacher prompt (Figure~\ref{fig:teacher-prompt-fb}). Unicode characters in the deployed prompt have been transliterated here for printability.}
\label{fig:critic-prompt}
\par\medskip
\endgroup

\subsection{Training configuration (full)}
\label{app:training_details}

This appendix lists complete hyperparameters for the self-distillation (with reference solution and ) and \grpo runs reported in the main paper. See Tables~\ref{tab:opsd-config}, \ref{tab:grpo-config}, \ref{tab:delta-from-zhao}, and \ref{tab:eval-config} for the values.

\begin{table}[ht]
\centering
\caption{Self-distillation training configuration. The step-aligned feedback block lists the additional critic-server settings.}
\label{tab:opsd-config}
\small
\begin{tabular}{lc}
\toprule
\textbf{Parameter} & \textbf{Value} \\
\midrule
\multicolumn{2}{l}{\emph{Model and adapters}} \\
Base model                       & Qwen3-1.7B \\
Precision                        & bfloat16 \\
Attention implementation         & Flash Attention 2 \\
LoRA rank $r$                    & 64 \\
LoRA $\alpha$                    & 128 \\
LoRA dropout                     & 0.0 \\
LoRA target modules              & \texttt{\{q,k,v,o,gate,up,down\}\_proj} \\
\midrule
\multicolumn{2}{l}{\emph{Optimization}} \\
Optimizer                        & AdamW \\
Learning rate                    & $5\!\times\!10^{-6}$ \\
LR schedule                      & constant \\
Weight decay                     & 0.0 \\
Gradient clipping (global)       & 0.1 \\
Per-device train batch size      & 2 \\
Gradient accumulation steps      & 4 \\
GPUs (data-parallel)             & 4 \\
\textbf{Effective batch size}    & \textbf{32} \\
Gradient checkpointing           & Enabled \\
DeepSpeed ZeRO stage             & 2 (CPU optimizer offload) \\
Epochs (max)                     & 7 \\
Save / eval interval             & every 10 optimizer steps \\
\midrule
\multicolumn{2}{l}{\emph{Sequence lengths}} \\
Max student completion $|\hat{y}|_{\max}$ & 2048 \\
Max packed sequence length       & 20{,}000 \\
\midrule
\multicolumn{2}{l}{\emph{On-policy sampling}} \\
Sampling backend                 & vLLM (colocate) \\
Temperature                      & 1.1 \\
Top-$p$                          & 0.95 \\
Top-$k$                          & 20 \\
Rollouts per problem             & 1 \\
Student thinking mode            & OFF \\
Teacher thinking mode            & ON \\
\midrule
\multicolumn{2}{l}{\emph{Distillation objective}} \\
Objective                        & Full-vocabulary generalized JSD \\
$\beta$ (forward KL)             & 0 \\
Per-token KL clip $\tau$         & disabled \\
Teacher policy                   & Fixed = initial (LoRA-disabled) base \\
Teacher temperature              & 1.0 \\
$\lambda$ (student-mixing)       & 1 \\
\midrule
\multicolumn{2}{l}{\stepfb\emph{ only: critic server}} \\
Critic model                     & Qwen/QwQ-32B (frozen) \\
Critic inference backend         & vLLM (separate node, OpenAI-compat HTTP) \\
Critic sampling                  & greedy ($T{=}0$), top-$p$ 0.95, top-$k$ off \\
Critic max output tokens         & 8{,}000 \\
Critic request timeout           & 240 s \\
Fallback on truncation           & Enabled \\
Reasoning-trace handling         & Strip everything up to and including \texttt{</think>} \\
\bottomrule
\end{tabular}
\end{table}

\begin{table}[ht]
\centering
\caption{\grpo baseline configuration \citep{shao2024deepseekmath}. Effective batch size and learning rate are matched to self-distillation experiments (\refsol and \stepfb); the larger generation cap and group size are the standard \grpo configuration.}
\label{tab:grpo-config}
\small
\begin{tabular}{lc}
\toprule
\textbf{Parameter} & \textbf{Value} \\
\midrule
Base model                       & Qwen3-1.7B \\
Learning rate                    & $5\!\times\!10^{-6}$ \\
Per-device batch size            & 1 \\
Gradient accumulation            & 8 \\
GPUs                             & 4 \\
Effective batch size             & 32 \\
LoRA $r$ / $\alpha$              & 64 / 128 \\
LoRA target modules              & \texttt{\{q,k,v,o,gate,up,down\}\_proj} \\
Max prompt length                & 2{,}048 \\
Max completion length            & 8{,}000 \\
Rollouts per prompt $G$          & 8 \\
Sampling temperature             & 1.2 \\
Loss type                        & \texttt{grpo} \\
Reward scaling                   & group-normalized \\
KL coefficient $\beta$           & 0 \\
PPO inner iterations             & 2 \\
Reward signal                    & binary, verified by \texttt{math\_verify} \\
\bottomrule
\end{tabular}
\end{table}

\begin{table}[ht]
\centering
\caption{Deltas of our setup from \citet{zhao2026opsd}. Items not listed are unchanged.}
\label{tab:delta-from-zhao}
\small
\begin{tabular}{lll}
\toprule
\textbf{Item} & \textbf{Zhao et al.} & \textbf{Ours} \\
\midrule
Training corpus       & OpenThoughts math subset & OpenMathReasoning (difficulty and format-filtered) \\
Teacher conditioning  & Reference solution (\refsol) & Critic feedback $f$ (\stepfb) \\
Student rollout cap   & 1{,}024 tokens & 2{,}048 tokens \\
Per-token KL clip $\tau$ & Enabled (untuned) & Disabled \\
Training budget       & 100 optimizer steps & Up to 7 epochs \\
\bottomrule
\end{tabular}
\end{table}

\begin{table}[ht]
\centering
\caption{Evaluation configuration.}
\label{tab:eval-config}
\small
\begin{tabular}{lc}
\toprule
\textbf{Parameter} & \textbf{Value} \\
\midrule
Inference backend           & vLLM \\
Tensor parallel size        & 4 \\
Samples per problem ($n$)   & 12 \\
Temperature                 & 1.0 \\
Top-$p$                     & 0.95 \\
Top-$k$                     & disabled \\
Max new tokens              & 38{,}912 \\
Thinking mode               & Enabled \\
Benchmarks                  & Left-out 30 problem test split from difficulty and format-filtered OpenMathReasoning \\
\bottomrule
\end{tabular}
\end{table}

\section{Sample solution, critique, and self-distillation advantage plots}
\label{app:samples}

\subsection{Example correct student solution}

\vspace{0.5cm}

\sample{Problem}{samples/ex1/problem.txt}
\sample{Reference Solution}{samples/ex1/reference.txt}
\sample{Student Solution}{samples/ex1/student.txt}
\sample{Step-level Critique}{samples/ex1/critique.txt}

\begin{figure}[h]
\centering
\includegraphics[width=\linewidth]{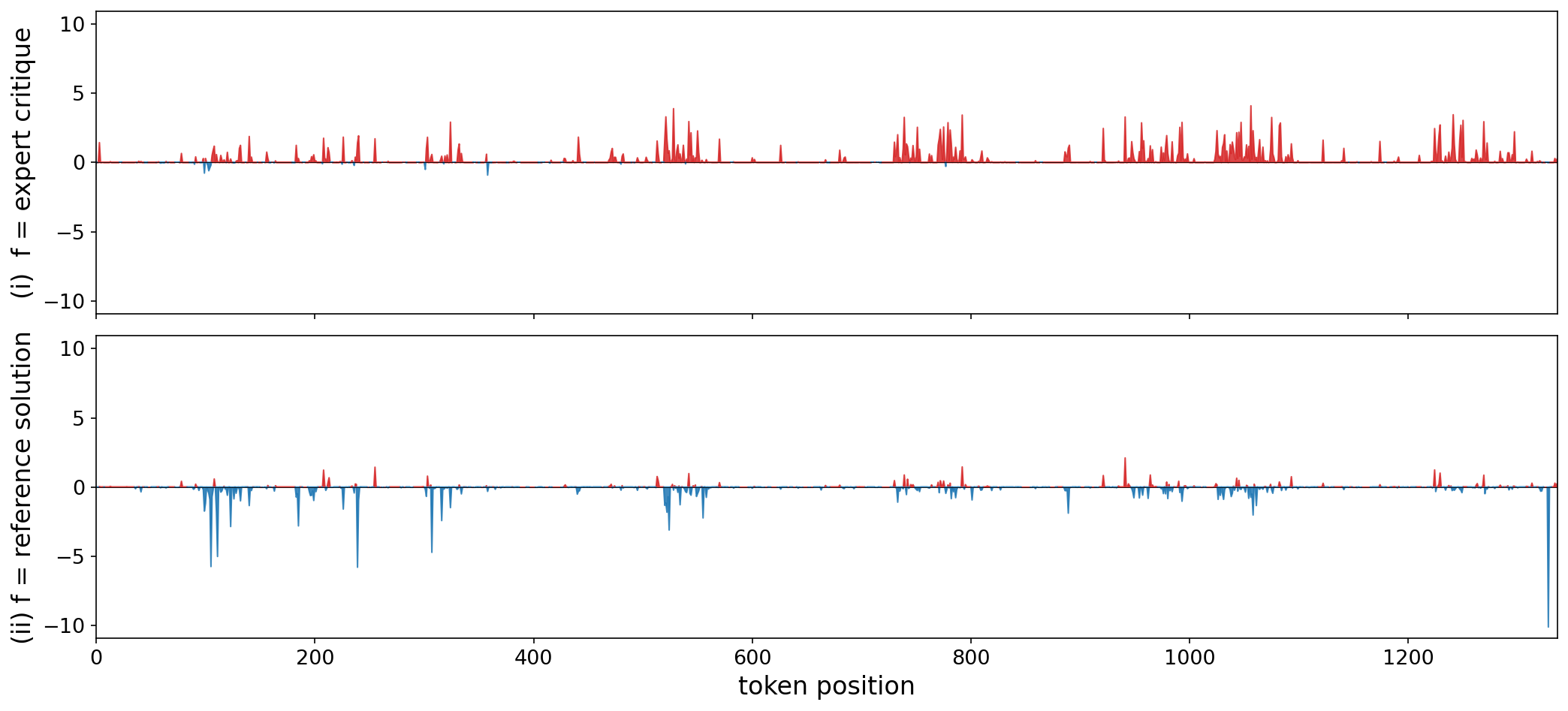}
\caption{Per-token advantages under each teacher context for correct student solution.}
\label{fig:adv_ex1}
\end{figure}

\subsection{Example incorrect student solution}

\vspace{0.5cm}

\sample{Problem}{samples/ex2/problem.txt}
\sample{Reference Solution}{samples/ex2/reference.txt}
\sample{Student Solution}{samples/ex2/student.txt}
\sample{Step-level Critique}{samples/ex2/critique.txt}

\begin{figure}[h]
\centering
\includegraphics[width=\linewidth]{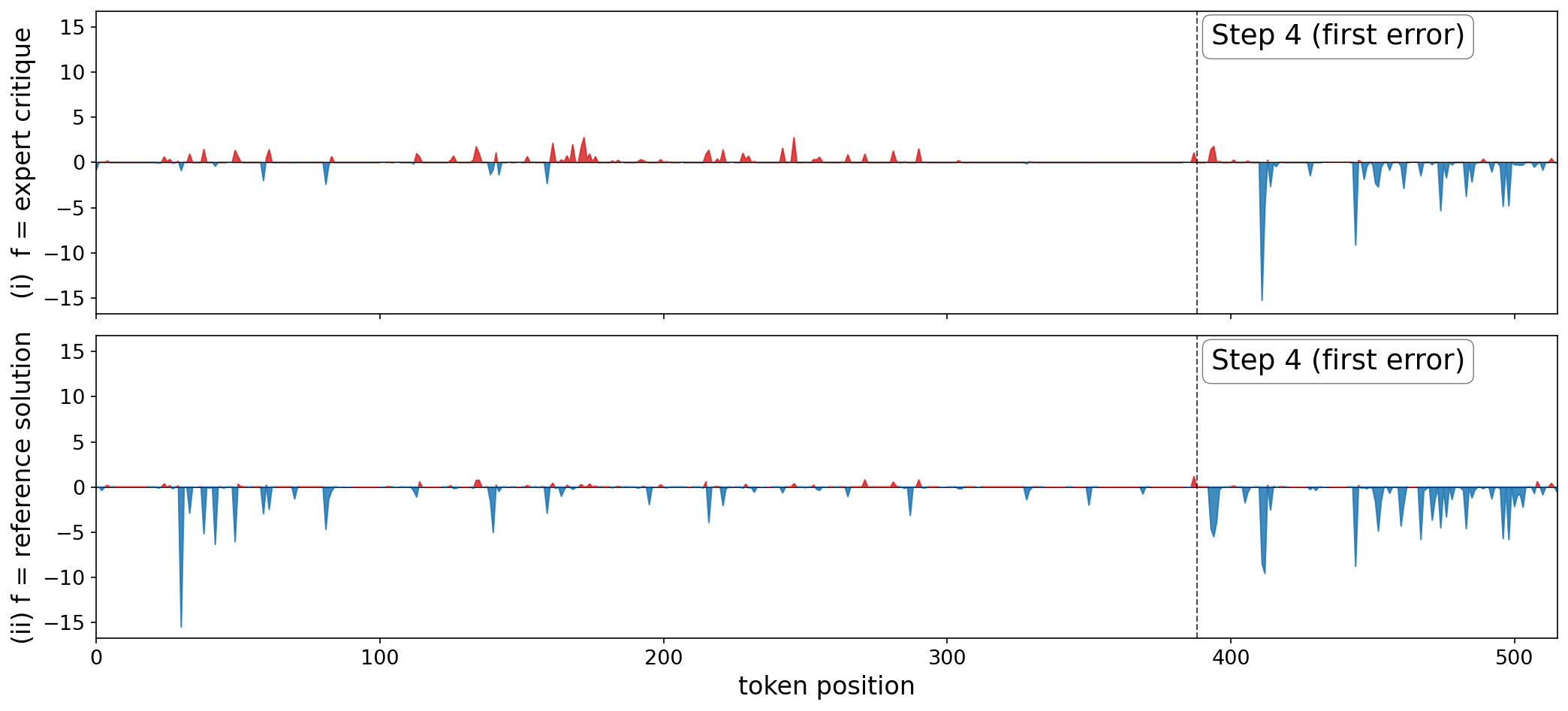}
\caption{Per-token advantages under each teacher context for incorrect student solution.}
\label{fig:adv_ex2}
\end{figure}


\end{document}